\documentclass[letterpaper, 10 pt, conference]{ieeeconf}  

\IEEEoverridecommandlockouts                              

\usepackage{graphics} 
\usepackage{amsmath} 
\usepackage{amssymb}  
\usepackage{graphicx}
\usepackage{algorithm}
\usepackage{algpseudocode}
\usepackage{arydshln}

\title{\LARGE \bf
SurgCalib: Gaussian Splatting-Based Hand-Eye Calibration for Robot-Assisted Minimally Invasive Surgery}

\author{Zijian Wu$^{1,*}$, Shuojue Yang$^{2,*}$, Yu Chung Lee$^{1}$, Eitan Prisman$^{3}$, Yueming Jin$^{2}$ and Septimiu E. Salcudean$^{1}$
\thanks{$^{*}$These authors contributed equally}
\thanks{$^{1}$Zijian Wu, Yu Chung Lee, and Septimiu E. Salcudean are with the Robotics and Control Laboratory,
        University of British Columbia, Vancouver, BC, Canada
        {\tt\small zijianwu@ece.ubc.ca}}%
\thanks{$^{2}$Shuojue Yang and Yueming Jin are with the Department of Biomedical Engineering, National University of Singapore, Singapore {\tt\small s.yang@u.nus.edu}}%
\thanks{$^{3}$Eitan Prisman is with the Division of Otolaryngology -- Head and Neck Surgery, University of British Columbia, Vancouver, BC, Canada}%
}

\begin{document}

\maketitle
\thispagestyle{empty}
\pagestyle{empty}

\begin{abstract}

We present a Gaussian Splatting-based framework for hand-eye calibration of the da Vinci surgical robot.
In a vision-guided robotic system, accurate estimation of the rigid transformation between the robot base and the camera frame is essential for reliable closed-loop control. 
For cable-driven surgical robots, this task faces unique challenges. The encoders of surgical instruments often produce inaccurate proprioceptive measurements due to cable stretch and backlash.
Conventional hand-eye calibration approaches typically rely on known fiducial patterns and solve the \textit{AX = XB} formulation. 
While effective, introducing additional markers into the operating room (OR) environment can violate sterility protocols and disrupt surgical workflows.
In this study, we propose SurgCalib, an automatic, markerless framework that has the potential to be used in the OR.
SurgCalib first initializes the pose of the surgical instrument using raw kinematic measurements and subsequently refines this pose through a two-phase optimization procedure under the RCM constraint within a Gaussian Splatting–based differentiable rendering pipeline.
We evaluate the proposed method on the public dVRK benchmark, SurgPose. The results demonstrate average 2D tool-tip reprojection errors of 12.24 px (2.06 mm) and 11.33 px (1.9 mm), and 3D tool-tip Euclidean distance errors of 5.98 mm and 4.75 mm, for the left and right instruments, respectively. 

\end{abstract}

 \section{INTRODUCTION}

In general robotics, hand-eye calibration is a fundamental procedure that establishes the 3D transformation between the robot base and the camera coordinate systems. 
This transformation enables measurements acquired in the camera frame to be mapped into the robot coordinate system for control and interaction.
In Robot-Assisted Minimally Invasive Surgery (RAMIS), e.g., da Vinci surgical system, articulated surgical instruments teleoperated as patient-side manipulators (PSM) and the endoscope, which provides a stereoscopic view of the surgical scene, inherently form a hand-eye system. 

Accurate hand-eye calibration can compensate for systematic biases in raw API-reported kinematic measurements and ensure robust surgical instrument end-effector pose tracking and tool tip localization, which serve as the foundation for downstream applications. For example, in a surgical Augmented Reality (AR) guidance system, accurate instrument tip positioning facilitates intraoperative registration between real-time ultrasound and the da Vinci robot~\cite{mohareri2015intraoperative}. This registration is an inevitable intermediate step for ultimately aligning multimodal preoperative medical imaging data (CT, MRI, and 3D Ultrasound) to the endoscopic video~\cite{kalia2021preclinical}. For autonomous surgical sub-task manipulation, such as suture needle grasping~\cite{zhong2020hand}, hand-eye calibration enables the detected needle pose in the camera frame to be transformed into the robot frame, thereby facilitating subsequent motion planning and control. More fundamentally, hand-eye calibration aligns the robot’s two essential sensing modalities, perception (vision) and proprioception, into a unified coordinate framework.

Despite its importance, hand-eye calibration on the da Vinci robotic system remains challenging due to inherent hardware limitations. 
The accuracy of the Cartesian end-effector pose cannot be arrived at solely from joint angle information, 
because of cable backlash and the structural flexibility of the tendon-driven instruments.
This issue is further exacerbated by the set-up joint (SUJ) controllers; since the SUJs lack active actuators and rely on passive encoders, if these are not read to update the kinematics in real time, the robot position is uncertain. In addition, there is a long and heavily loaded kinematic chain from the setup arms to the camera of the system, leading to unreliable instrument tip positioning~\cite{cui2023caveats}. Prior efforts have attempted to mitigate these inaccuracies using data-driven or marker-based approaches. For instance, Hwang \textit{et al.}~\cite{hwang2020efficiently} propose a method using the RGB-D fiducial markers and a recurrent neural network. Yet, the reliance on a specialized depth camera and the time-consuming model training limit its practicality. Similarly, a hierarchical scheme introduced by Lu \textit{et al.}~\cite{lu2022unified} requires manual labeling of instrument features and specific controlled motions for local refinement. Such dependencies not only hinder full system automation but also significantly increase preoperative setup time, impairing their practicability in clinical applications. 

Recent advances in robot pose estimation demonstrate significant potential to replace marker-based pose estimation and tracking approaches~\cite{greene2025markerless,lee2020camera}. Deep learning-based feature detection methods enable reliable 2D keypoint localization of surgical instruments directly from visual input~\cite{wu2025surgpose,ghanekar2025video,wu2025tooltipnet}, providing essential geometric cues for markerless pose estimation~\cite{liang2025efficient,lu2023markerless}. Furthermore, leveraging differentiable rendering techniques, prior work~\cite{yang2025instrument, liang2025differentiable} performs single-frame robot pose estimation in a render-and-compare paradigm, where the robot pose is iteratively refined by minimizing discrepancies between rendered silhouettes and observed segmentation masks. 
More recently, with the emergence of neural rendering, particularly Gaussian Splatting (GS)~\cite{kerbl20233d}, a high-fidelity and differentiable robot representation~\cite{yang2025instrument}, one may incorporate photometric consistency between rendered textures and real images, leading to improved pose estimation accuracy. 
A distinctive characteristic of RAMIS robot arms is their Remote Center of Motion (RCM):  surgical instruments are inserted into the patient’s body through small incisions, and the motions of the instrument shaft are constrained to pivot around these fixed entry points.  However, existing single-frame pose estimation methods typically neglect this physical constraint. Consequently, the estimated shaft axes may fail to satisfy the RCM geometry, resulting in kinematically inconsistent solutions and degraded pose accuracy. 

In this work, we propose SurgCalib, a fully automatic and markerless hand-eye calibration framework for the da Vinci surgical system. SurgCalib first estimates the initial pose of the surgical instrument using a set of 2D-3D correspondences. The 2D keypoints are detected by a deep learning-based keypoint detector, and the 3D keypoints are derived from forward kinematics given raw joint angles. Then we integrate 3D GS-based instrument representation within a differentiable rendering pipeline to refine the surgical instrument pose in a render-and-compare manner, without the need for manual annotation or fiducial markers. By adopting a two-phase optimization strategy that explicitly incorporates the RCM constraint, our framework effectively estimates instrument poses across a sequence while preserving geometric consistency. Finally, the hand-eye transformation is computed by solving a least-squares optimization problem given the refined pose estimates. 

This approach achieves promising accuracy while requiring minimal input, namely, a monocular video of arbitrary instrument motion and the corresponding kinematic data from the control software. The main contributions of this work are summarized as follows:
\begin{itemize}
  \item We develop an automatic hand-eye calibration pipeline that requires only random instrument motion captured by a monocular endoscopic camera and raw kinematic measurements, eliminating the need for manual feature labeling or carefully designed calibration trajectories.
  \item We present the first application of 3D Gaussian Splatting to the surgical robot hand-eye calibration problem, leveraging its high-fidelity and differentiable rendering for robust pose optimization.
  \item We propose a RCM-aware, two-phase pose optimization strategy that progressively refines the RCM position and instrument poses across a sequence, enforcing geometric consistency and compensating for kinematic uncertainties.
  \item Finally, we quantitatively evaluate the proposed framework on the public dVRK benchmark, SurgPose. We report both 2D reprojection errors and 3D tool-tip localization errors after hand-eye calibration. 
\end{itemize}
Section II presents related work, our problem is formulated in Section III, solution methods in Section IV, quantitative results in Section V and a discussion in Section VI.
Qualitative results and visualizations are provided in the accompanying supplementary video.

\section{RELATED WORK}
Hand-eye calibration has been a long-standing problem in robotics. Existing approaches can be categorized into two classes: 1) classic approaches; 2) learning-based approaches. In this study, we place particular emphasis on hand-eye calibration methods customized for RAMIS applications.

\textit{Classic Hand-eye Calibration:} Since the 1980s, studies on hand-eye calibration have centered on solving the well-known formulation $AX=XB$, first proposed by Shiu and Ahmad~\cite{88014}. In this formulation, $A$ and $B$ denote the motion of the robot end-effector and the camera, while $X$ represents the hand-eye transformation to be estimated. Early  analytical methods, e.g.,~\cite{tsai1989new}, decouple rotation and translation for closed-form estimation. Various mathematical representations, e.g., Lie groups~\cite{park1994robot} and quaternions~\cite{daniilidis1999hand}, have been proposed to improve numerical compactness and stability. All above-mentioned methods treat hand-eye calibration as a sensor-agnostic rigid-body motion problem, without incorporating modality-specific geometric objectives (e.g., reprojection error in vision systems), potentially limiting achievable accuracy. 

\textit{Learning-Based Hand-Eye Calibration:} With the advancements of deep learning, many components in the hand-eye calibration pipeline have been revolutionized by learning based techniques. Lee \textit{et al.}~\cite{lee2020camera} present a method that uses a deep neural network (DNN) to detect keypoints of the robot and then uses Perspective-n-point (PnP) to estimate the camera-to-robot transformation. Labbé \textit{et al.}~\cite{labbe2021single} propose a camera-to-robot pose estimation model in a render-and-compare manner requiring only a single view. Lu \textit{et al.}~\cite{lu2023markerless} further improve this line of work by introducing a backpropagatable PnP solver. Tang \textit{et al.}~\cite{tang2025kalib} propose an easy-to-setup approach with two basic prerequisites, the robot’s kinematic chain and a predefined reference point on the robot, based on the point tracking foundation model. These methods are primarily designed for general-purpose robotic manipulators, which typically feature rigid structures, large workspaces, and accurate proprioceptive sensing. Consequently, they cannot be directly transferred to surgical robotic systems.   

\textit{Hand-Eye Calibration for Surgical Robots:} Early studies on hand-eye calibration for surgical robots primarily adopted classical numerical or analytical methods. Pachtrachai \textit{et al.}~\cite{pachtrachai2016hand} and Zhang \textit{et al.}~\cite{zhang2017computationally} both employ a dual-quaternion representation of the hand-eye rigid transformation. The former introduces a customized calibration pattern that can be grasped by the surgical instrument, whereas the latter proposes a calibration object-free method. Hwang \textit{et al.}~\cite{hwang2020efficiently}, and Pachtrachai \textit{et al.}~\cite{pachtrachai2021learning} adopt DNNs to predict the hand-eye matrix. Compared to methods that recover a fixed matrix, these learning-based approaches can model dynamic kinematic errors and compensate for time-varying system inaccuracies. Zhong \textit{et al.}~\cite{zhong2020hand} propose a method that leverages interactive manipulation of the instrument without requiring visual feature detection. In vision systems where focus changes often, e.g., the da Vinci Si surgical system, the camera matrix varies with depth. To adapt this changing case, Kalia \textit{et al.}~\cite{kalia2021preclinical} jointly estimate the hand-eye and camera calibration by minimizing the keypoint reprojection loss. Recently, Cui \textit{et al.}~\cite{cui2026fly} propose an online hand-eye calibration framework based on a training-free keypoint association algorithm using analytical Jacobian matrices. 

\section{TASK FORMULATION} 
\begin{figure}[t]
\centering
\includegraphics[width=\columnwidth]{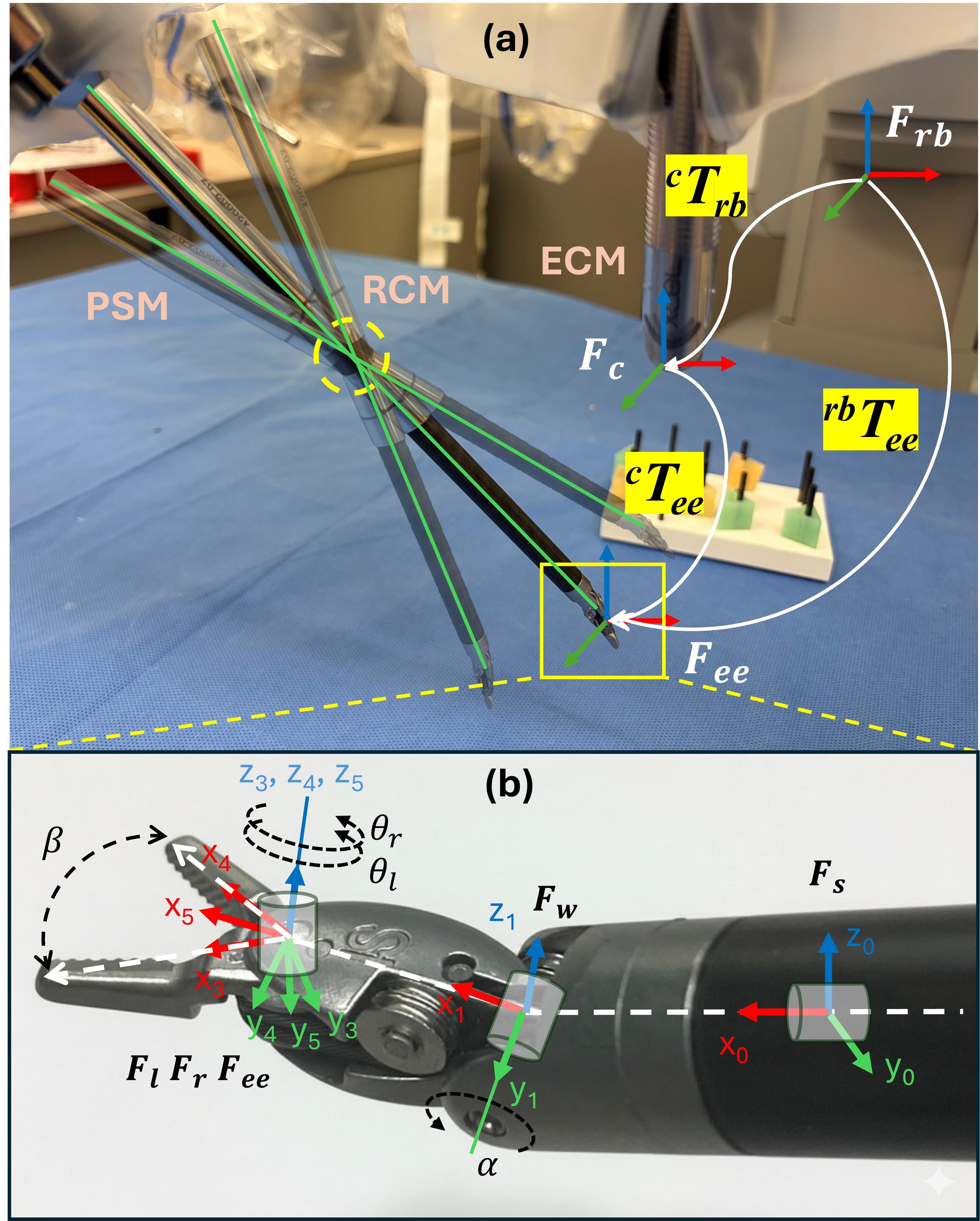}
\caption{(a) The diagram of task definition. The RCM point lies within the yellow dashed circle. The shaft centerlines are shown as green lines. (b) The frame definition of the EndoWrist surgical instrument.}
\label{fig1}
\end{figure}
As depicted in Fig.~\ref{fig1}(a), let $\mathcal{F}_{c}$ and $\mathcal{F}_{ee}$ denote the coordinate systems of the left endoscopic camera and the surgical instrument's end-effector, respectively. We define $\mathbf{p}_{rcm}\in\mathbb{R}^3$ as the position of the Remote Center of Motion (RCM) with respect to (w.r.t.) the camera frame $\mathcal{F}_{c}$.

The objective of this hand-eye calibration is to determine the camera to robot base transformation, ${}^{c}\mathbf{T}_{rb}\in SE(3)$, which transforms the end-effector pose ${}^{rb}\mathbf{T}_{ee}$ w.r.t. the robot base frame $\mathcal{F}_{rb}$ to the pose ${}^{c}\mathbf{T}_{ee}$ w.r.t. camera frame $\mathcal{F}_{c}$. This relationship is formulated as:
\begin{equation} 
{}^{c}\mathbf{T}_{ee} = {}^{c}\mathbf{T}_{rb} \cdot {}^{rb}\mathbf{T}_{ee}, 
\end{equation}
where all transformation matrices belong to the $SE(3)$.
The raw pose ${}^{rb}\mathbf{T}_{ee}$ can be obtained via open-source control software, e.g., da Vinci Research Kit (dVRK)~\cite{kazanzides2014open}, or the built-in research API (dVAPI) of the da Vinci robotic system. The specific definition of the robot base frame varies across robot versions and control software.

Following \cite{yang2025instrument}, the forward kinematics of the Large Needle Driver (LND) is defined in Fig. \ref{fig1}(b). The coordinate frames for the shaft, wrist, and the left and right grippers are denoted by $\mathcal{F}_{s}$, $\mathcal{F}_{w}$, $\mathcal{F}_{l}$, and $\mathcal{F}_{r}$, respectively. Notably, $\mathcal{F}_{l}$, $\mathcal{F}_{r}$, and the end-effector frame $\mathcal{F}_{ee}$ share a common origin and a coincident $z$-axis. Their rotational relationship is defined such that the $x$-axis of $\mathcal{F}_{ee}$ bisects the gripper jaw angle. This is expressed as:
\begin{equation}
\beta = \frac{\theta_{l} + \theta_{r}}{2},
\end{equation}
where $\theta_{l} \geq \beta \geq \theta_{r}$ to ensure the definition is consistent with the mechanical constraints of the instrument.
 
The articulation and global pose of the surgical instrument w.r.t. $\mathcal{F}_{c}$ is represented by the set $\mathbf{q}=\{\theta_l, \theta_r, \alpha, {}^{c}\mathbf{T}_s\}$. 
Based on the forward knematics, the corrected pose ${}^{c}\mathbf{T}_{ee}$ can be computed as:
\begin{equation}
{}^{c}\mathbf{T}_{ee}={}^{c}\mathbf{T}_{s}\cdot{}^{s}\mathbf{T}_{w}\cdot{}^{w}\mathbf{T}_{l}\cdot{}^{l}\mathbf{T}_{ee},
\end{equation}
where each transformation $\mathbf{T} \in SE(3)$.

Let the centerline of the instrument's shaft be denoted by the line $\mathcal{L}_{s}$. 
The origin of $\mathcal{F}_s$, denoted by $\mathbf{o}_s \in \mathbb{R}^3$, is a point lying on $\mathcal{L}_{s}$. The unit vector aligned with the shaft's longitudinal axis is assigned to the $x$-axis of $\mathcal{F}_s$, denoted by $\mathbf{x}_s$. Consequently, any point $\mathbf{p}\in\mathbb{R}^3$ on the shaft axis can be parameterized as:
\begin{equation} \mathbf{p}(\gamma) = \mathbf{o}_s + \gamma \cdot\mathbf{x}_s, \end{equation}
where $\gamma \in \mathbb{R}$ is a scalar parameter representing the signed distance from the origin $\mathbf{o}_s$ along the axis.

\begin{figure*}[t]
\centering
\includegraphics[width=\textwidth]{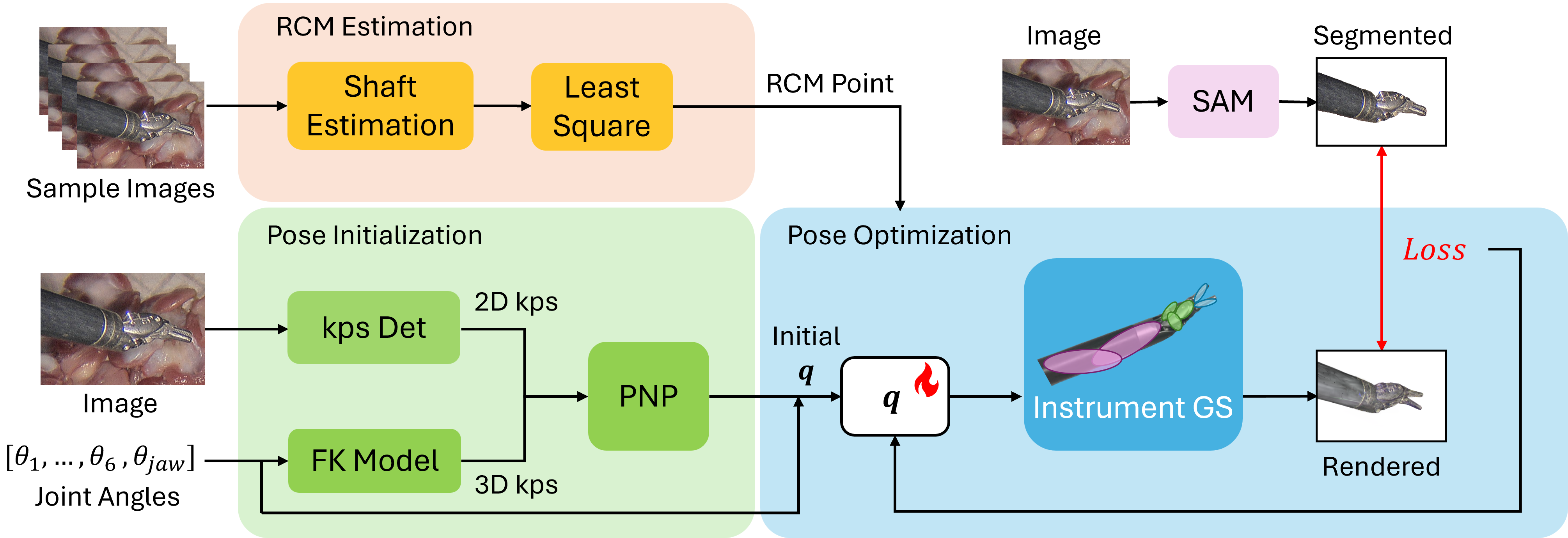}
\caption{The schematic  of our proposed pose initialization and refinement method.}
\label{fig2}
\end{figure*}

\section{METHODS}
\subsection{GS Representation of Surgical Instruments} 
With the emergence of neural rendering, GS-based robot representations~\cite{liu2024differentiable} combine the merits of both controllability and high-fidelity textures. With a fully differentiable pipeline, these representations allow for the optimization of robot states, e.g., joint angles and end-effector poses, directly through image-based losses. In this work, we adopt Instrument-Splatting~\cite{yang2025instrument} to represent the surgical instrument as an articulated collection of 3D Gaussians. Unlike traditional mesh-based rendering, this representation achieves photorealistic rendering while maintaining the computational efficiency required for iterative optimization.

Each Gaussian is defined by its position $\boldsymbol{\mu}\in\mathbb{R}^{3}$, a covariance matrix $\boldsymbol{\Sigma}_j$ (scaling vector $\mathbf{s} \in \mathbb{R}^3$ and a quaternion $\mathbf{r} \in \mathbb{R}^4$), opacity $\boldsymbol{\alpha}\in\mathbb{R}$, and spherical harmonic coefficients $\boldsymbol{sh}\in\mathbb{R}^{27}$ for view-dependent appearance. The surgical instrument is represented as 3D Gaussians that are partitioned into semantic sets $\mathcal{G}_k$, each bound to a rigid part $k \in \{s, w, l, r\}$, corresponding to the shaft, wrist, and left/right grippers, respectively. For a Gaussian point $\mathbf{j}$ of part $k$, its position $\boldsymbol{\mu}'_j$ and rotation $\mathbf{r}'_j$ w.r.t. the camera frame are computed via the kinematic chain:
\begin{equation}
\boldsymbol{\mu}'_j = {}^{c}\mathbf{T}_k \cdot \boldsymbol{\mu}_j;\;\;\boldsymbol{r}'_j = {}^{c}\mathbf{R}_k \cdot \boldsymbol{r}_j,
\end{equation}
in which
\begin{equation}
{}^{c}\mathbf{T}_k =
\begin{cases}
{}^{c}\mathbf{T}_s, 
& k = s \\[4pt]
{}^{c}\mathbf{T}_s\cdot{}^{s}\mathbf{T}_w, 
& k = w \\[4pt]
{}^{c}\mathbf{T}_s\cdot{}^{s}\mathbf{T}_w\cdot{}^{w}\mathbf{T}_{l/r}, 
& k = l/r
\end{cases}
\end{equation}
where ${}^{s}\mathbf{T}_w$ and ${}^{w}\mathbf{T}_k$ are local transformations derived from the joint angles, ${}^{c}\mathbf{R}_k$ denotes the rotational component of rigid transformation ${}^{c}\mathbf{T}_k$.

\begin{figure}[t]
\centering
\includegraphics[width=\columnwidth]{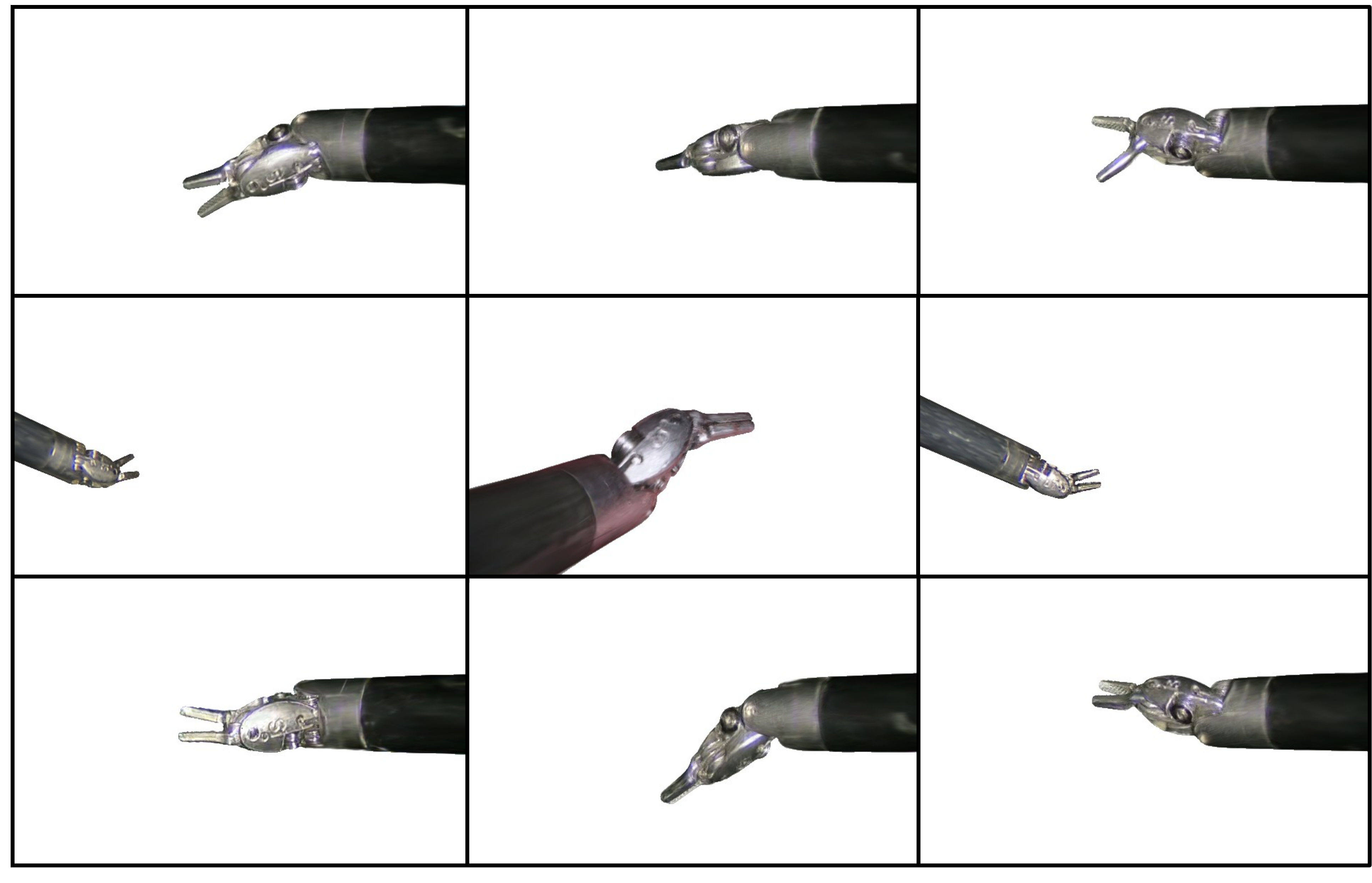}
\caption{Example images rendered by Instrument-Splatting.}
\label{fig3}
\end{figure}

As shown in Fig.~\ref{fig3}, Instrument-Splatting enables the rendering of a Large Needle Driver (LND) with high visual fidelity on arbitrary poses. Using differentiable rasterization, the gradients from a visual-wise loss can be back-propagated through the pipeline. This makes the representation uniquely suited for our two-phase optimization strategy, as it provides a dense, texture-aware signal that is more robust than sparse keypoint matching alone.

\subsection{Visual Features Extraction} 
With advances in computer vision, segmentation~\cite{wu2025augmenting,yue2024surgicalsam} and keypoint detection models~\cite{han2025robust} have achieved significant performance improvements. In the proposed framework, we integrate two visual perception modules, Instance Segmentation and Keypoint Detection, to provide necessary visual cues for pose initialization and subsequent pose optimization. For instance segmentation, we utilize Segment Anything Model 2~\cite{ravi2024sam} (SAM 2) to segment the surgical instrument. SAM 2 can robustly generate accurate instance masks of surgical instruments in the dry-lab environment without any task-specific fine-tuning.

For keypoint detection, we adopt a deep learning-based approach, MFC-tracker~\cite{ghanekar2025video}. This model is based on supervised deep learning architectures, e.g., DeepLabV3~\cite{chen2017rethinking}, adding an additional refinement model for robust tracking. We choose the SurgPose~\cite{wu2025surgpose} videos 5-7, 30-33 as the training set. 
All training images are resized to 640$\times$512 and perform the default data augmentation strategy. The training is conducted for 200 epochs on a single NVIDIA RTX 3090Ti GPU.

\subsection{Pose Initialization} Given a monocular image and its corresponding raw joint angles, this pose initialization module estimates a coarse initial pose of the surgical instrument. Optimization-based pose estimation methods, such as the render-and-compare approach, require a computationally feasible initial estimate to ensure convergence and avoid local minima. We utilize noisy joint angle readings from the dVRK or research API to configure the instrument's kinematic model. From this configuration, the 3D positions of the instrument's keypoints are derived via forward kinematics.

While these 3D keypoints are subject to noisy joint angles, they provide sufficient spatial constraints for a PnP solver. We leverage the keypoint detector introduced in the previous section to extract the corresponding 2D keypoints from the image. Using these 2D-3D correspondences, we adopt the EPnP~\cite{lepetit2009ep} algorithm to solve the initial pose $\mathbf{q}_{init}$.

\subsection{RCM Estimation} 
The remote center of motion (RCM) is a key design in robot-assisted minimally invasive surgery (MIS) systems, as surgical instruments must pivot around a fixed incision point on the patient's body to minimize trauma.
In practice, due to mechanical tolerances and joint misalignments, the RCM is often not a singular point but a localized region (frequently modeled as a sphere~\cite{lu2022unified}) during motion. Given a sequence of $N$ observations, we denote the estimated shaft centerlines as $\{\mathcal{L}_{s,n}\}_{n=1}^N$. The RCM $\mathbf{p}_{rcm}$ is formulated as the point that minimizes the sum of squared perpendicular distances to these lines: 
\begin{equation}
\mathbf{p}_{rcm} = \underset{\mathbf p}{\arg\min}\sum_{n=1}^N\left\|dist(\mathbf p,\;\mathcal{L}_{s,n})\right\|^2\\
\end{equation}
where the distance function $dist(\cdot)$ is defined using the orthogonal projection onto the line:
\begin{equation}
    dist(\mathbf{p}, \mathcal{L}_{s,n}) = (\mathbf{I} - \mathbf{x}_{s,n} \mathbf{x}_{s,n}^\top)(\mathbf{p} - \mathbf{o}_{s,n}).
\end{equation}
This formulation is a linear least-squares problem, which can be solved efficiently in closed form via the method of least-squares intersection. According to equations (5) and (6), we can compute $\mathbf{p}_{rcm,init}$ given $\{\mathcal{L}_{s,init,n}\}_{n=1}^N$.

\subsection{Two-Phase Pose Optimization}

To obtain accurate poses $\{\mathbf{q}_{opt,n}\}_{n=1}^N$, we perform a two-phase optimization initialized from $\{\mathbf{q}_{init,n}\}_{n=1}^N$. 
Since the initial RCM position $\mathbf{p}_{rcm,init}$ is estimated from noisy shaft centerlines, it does not reliably represent the true RCM point. 
To progressively enforce geometric consistency while avoiding early over-constraint, we design a two-phase optimization strategy, which is illustrated in Algorithm 1.
    
    

\textbf{Phase 1 (Global RCM Refinement):} We jointly optimize the pose parameters while dynamically updating the RCM position per epoch. 
This phase runs for a fixed number of $M$ epochs. 
At the end of each epoch, the RCM position $\mathbf{p}_{rcm}$ is recomputed based on the updated pose estimates to progressively refine the RCM geometric constraint. 
The loss is defined as 
\begin{equation}
    L_{phase1}=\lambda_s L_{silh}+\lambda_p L_{px}+\lambda_k L_{kpt},
\end{equation}
where $L_{silh}$ and $L_{px}$ denote the $L_1$ losses between the rendering and the segmented instrument; $L_{kpt}$ is the Chamfer loss of keypoints.
Note that we exclude the RCM loss in this phase. 
As the RCM position is still being refined, prematurely enforcing the RCM constraint could steer the optimization toward an incorrect kinematic configuration and hinder convergence.

Since shaft centerline extraction may be noisy, directly fitting the RCM using all lines can lead to biased estimates. 
To address this, we introduce an iterative outlier rejection least-squares procedure. 
Specifically, after estimating the RCM, shaft centerlines with large orthogonal residuals are removed, and the RCM is recomputed using only inlier lines. 
This robust estimation step mitigates the influence of erroneous shaft axes observations and improves global geometric consistency.

\textbf{Phase 2 (Per-Frame Pose Refinement with RCM Constraint):} 
After Phase 1, the RCM position is frozen and treated as a fixed geometric constraint. 
We then perform a single-frame pose refinement for each frame independently. 
For image $\mathcal{I}_n$, the render-and-compare optimization runs iteratively, and is terminated early if the total loss does not decrease for $K$ consecutive iterations. 
The loss function in Phase 2 is defined as: 
\begin{equation}
    L_{phase2}=\lambda_s L_{silh}+\lambda_p L_{px}+\lambda_k L_{kpt}+\lambda_r L_{rcm}. 
\end{equation}
As formulated in equation (8), RCM Loss $L_{rcm}$ is the mean squared orthogonal distance, which can be denoted as:
\begin{equation}
L_{rcm}
=
\frac1N\sum_{n=1}^{N} dist(\mathbf{p}_{rcm}, \mathcal{L}_{s,n}) 
\end{equation}

After this two-phase pose optimization, we have a set of refined poses $\{\mathbf{q}_{opt}\}_{n=1}^{N}$ of the surgical instrument end-effector w.r.t. the camera frame $\mathcal{F}_{c}$.

\subsection{Compute Compensation Transformation}
Given $\{({}^{c}\mathbf{T}_{ee,n}, {}^{rb}\mathbf{T}_{ee,n})\}_{n=1}^N$, $N$ pairs of end-effector poses in $\mathcal{F}_c$ and $\mathcal{F}_{rb}$, we aim to determine the optimal rigid transformation ${}^{c}\mathbf{T}_{rb}$ that minimizes the discrepancy between the end-effector poses in these two frames. 
To solve this in a least-squares sense, we extract their translational components (two sets of points) ${\{{\mathbf p}_{c,n}\}}_{n=1}^N$ and $\{\mathbf{p}_{rb,n}\}_{n=1}^N$.
The transformation ${}^{c}\mathbf{T}_{rb}$ can be solved using the Kabsch-Umeyama algorithm~\cite{umeyama2002least}, which finds the optimal rotation $\mathbf{R}^\star$ and translation $\mathbf{t}^{\star}$ by minimizing the root-mean-square deviation (RMSD) between the two point clouds:
\begin{equation}
\mathbf{R}^\star,\mathbf{t}^\star =\arg\min_{\mathbf{R}, \mathbf{t}} \sum_{n=1}^N w_n| \mathbf{p}_{c,n} - (\mathbf{R}\cdot \mathbf{p}_{rb,n} + \mathbf{t}) |^2,
\end{equation} 
The optimal transformation matrix is ${}^{c}\mathbf{T}_{rb}=[\mathbf{R}^\star|\mathbf{t}^\star]$.

\begin{figure*}[h]
\centering
\includegraphics[width=\textwidth]{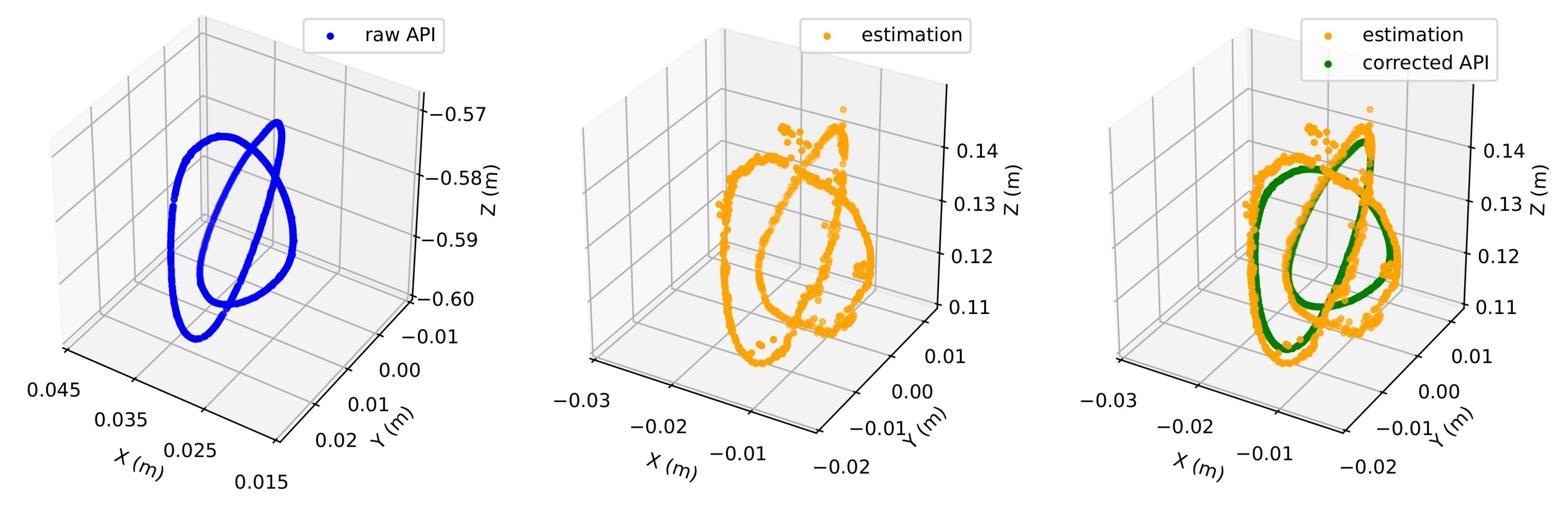}
\caption{The trajectories of end-effector origins. From left to right: raw data from the dVRK, estimated positions by the proposed pose refinement approach, and the aligned trajectories using hand-eye transformation.}
\label{fig_4}
\end{figure*}

\begin{figure*}[t]
\centering
\includegraphics[width=\textwidth]{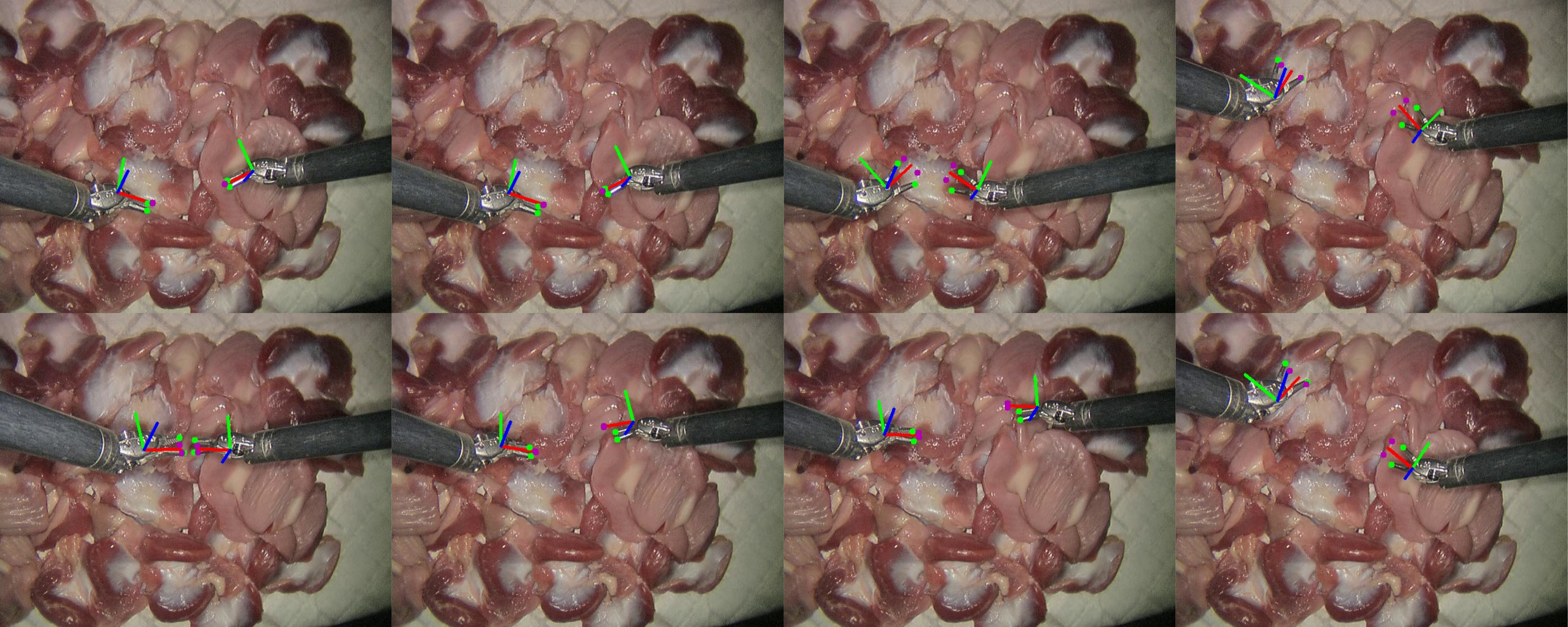}
\caption{The visualization of the compensated end-effector pose. The green and purple dots are the ground truth and reprojected tool tips, respectively. The upper and bottom row refers to frames with fewer and larger errors.}
\label{vis}
\end{figure*}

\begin{algorithm}[t]
\small
\caption{Two-Phase Pose Optimization}
\begin{algorithmic}[1]

\Procedure{OptimizePose}{$\{\mathbf{q}_{init,n}\}_{n=1}^N, \{\mathcal{I}_n\}_{n=1}^N$}
    \State $\{\mathbf{q}_{n}\}_{n=1}^N \gets \{\mathbf{q}_{init,n}\}_{n=1}^N$

    \State \textbf{Phase 1: Global Refinement}
    \For{$epoch = 1$ to $M$}
        \State Update $\{\mathbf{q}_{n}\}_{n=1}^N$
        \State Estimate $\mathbf{p}_{rcm}$
        \State Perform outlier rejection and re-estimate $\mathbf{p}_{rcm}$
    \EndFor

    \State Freeze $\mathbf{p}_{rcm}$

    \State \textbf{Phase 2: Per-Frame Refinement}
    \For{$n = 1$ to $N$}
        \While{loss decreases within $K$ iterations}
            \State Update $\mathbf{q}_{n}$
        \EndWhile
        \State $\mathbf{q}_{opt,n} \gets \mathbf{q}_n$
    \EndFor

    \State \Return $\{\mathbf{q}_{opt,n}\}, \mathbf{p}_{rcm}$

\EndProcedure
\end{algorithmic}
\end{algorithm}
\section{EVALUATION \& RESULTS}
\begin{figure}[t]
\centering
\includegraphics[width=\columnwidth]{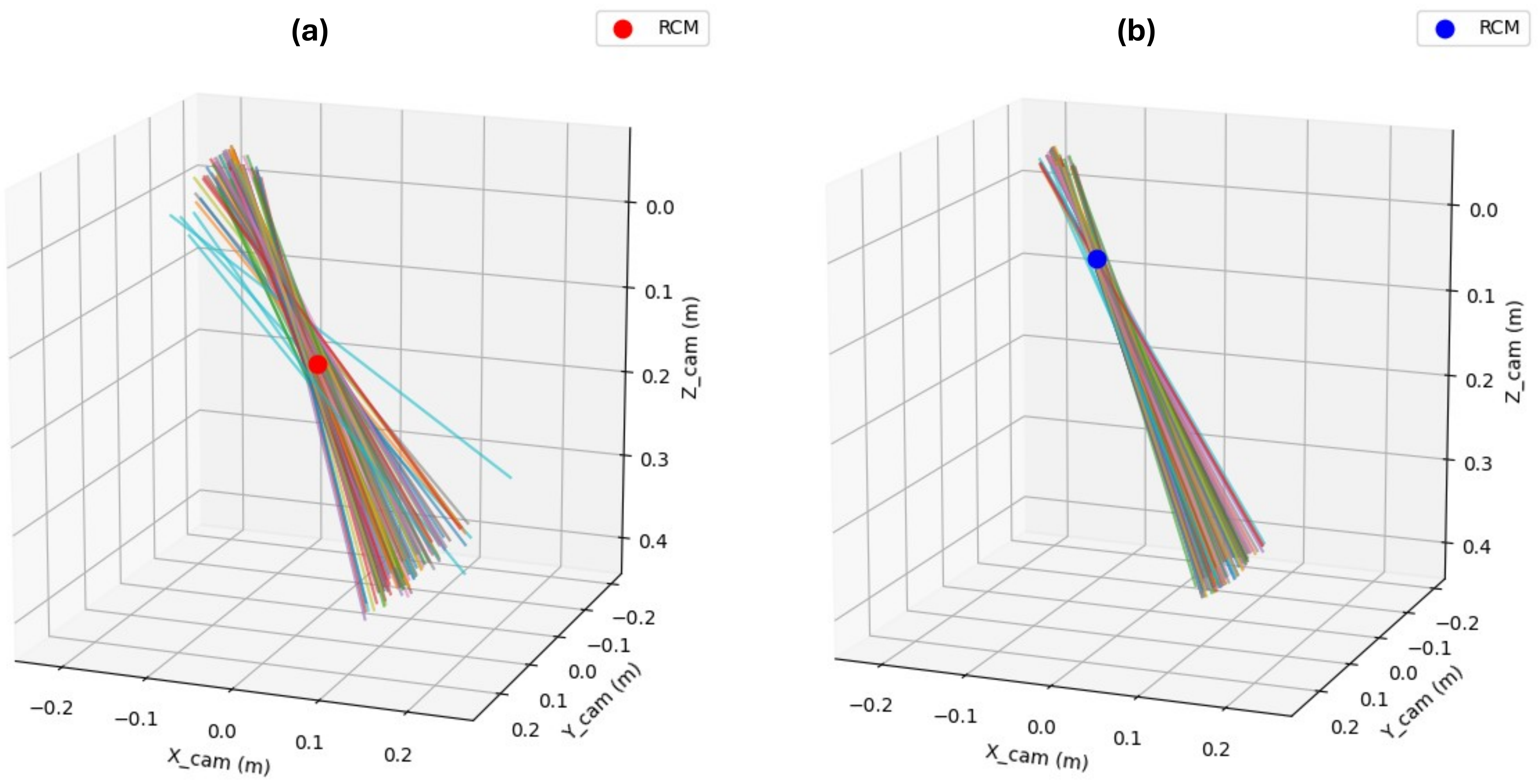}
\caption{Visual comparison of the shaft centerline convergence around RCM. (a) Shaft axes after phase 1 optimization; (b) Shaft axes after phase 2 optimization with the RCM constrain. Note that the RCM (blue point) is re-estimated after optimization.}
\label{fig_rcm}
\end{figure}

To evaluate the accuracy and effectiveness of our methods, we perform experiments on a public dataset.
We adopt SurgPose~\cite{wu2025surgpose}, which is collected on the first-generation da Vinci robotic system, as the public benchmark to evaluate the performance of SurgCalib on the dVRK platform. SurgPose provides binocular videos, stereo camera calibration, annotation of 2D keypoints, and the associated kinematic data (7D joint angles and 6D end-effector poses) read from dVRK. In this study, we resize all frames to size 640$\times$512. Specifically, we use the video 0 to 4 (1000 frames per video) of SurgPose for experiments. We take video 0 as the training data to estimate the hand-eye transformation ${}^{c}\hat{ \mathbf{T}}_{rb}$. We directly use this ${}^{c}\hat{ \mathbf{T}}_{rb}$ to correct the kinematic data of Videos 1-4 and compute the metrics without any training. According to the previous problem formulation, we are solving the least-squares problem: 
\begin{equation}
        {}^{c}\hat{ \mathbf{T}}_{ee}={}^{c}\hat{ \mathbf{T}}_{rb}\cdot{}^{rb}\mathbf{T}_{ee}^\#.
\end{equation}
SurgPose provides the $\{{}^{rb}\mathbf{T}_{ee}^\#\}_{n=1}^{1000}$ reported by the dVRK. 
For the data sequence in Video 0, the proposed method estimated a set of poses $\{{}^{c}\hat{\mathbf{T}}_{ee,n}\}_{n=1}^{N}$, where $N$ denotes the number of valid frames after excluding outlier poses. We then use the Kabsch-Umeyama algorithm to recover the ${}^{c}\hat{ \mathbf{T}}_{rb}$.
 Fig.~\ref{fig_4} is a visual comparison of compensated dVRK trajectories and estimated trajectories using the proposed pose refine strategy. Fig.~\ref{vis} is the visualization of the end-effector pose after hand-eye calibration. We visualize the shaft centerlines before and after adding RCM loss in Fig~\ref{fig_rcm} to qualitatively demonstrate that the RCM constraint can effectively ensure that the poses satisfy this physical constraint.

\begin{figure}[h]
\centering
\includegraphics[width=\columnwidth]{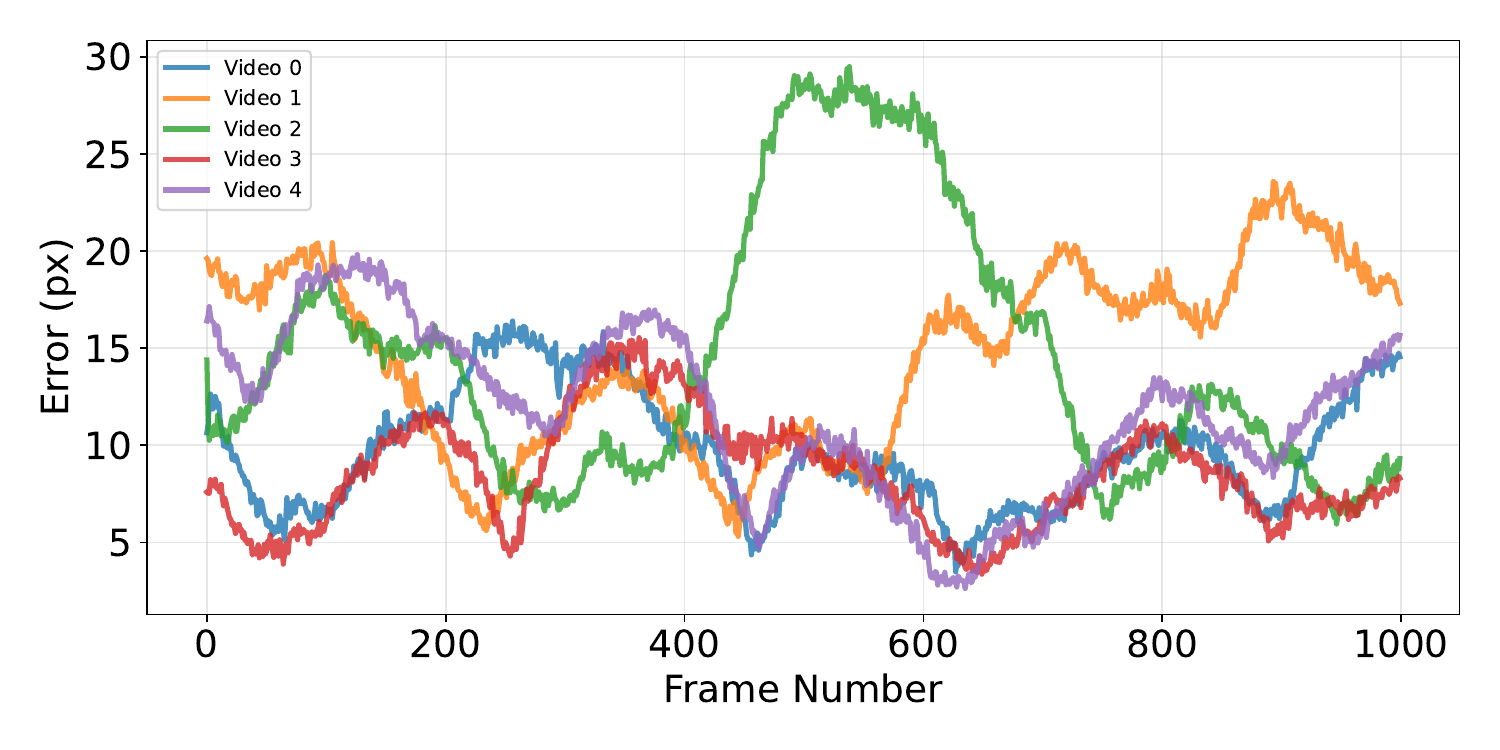}
\caption{The tool tip reprojection error distribution across the frame numbers.}
\label{fig_error}
\end{figure}

To quantitatively validate this ${}^{c}\hat{ \mathbf{T}}_{rb}$, we project the positions of tool tips to the image given the camera matrix. We separately do the hand-eye calibration for the left and right instruments. We employ the average and median Euclidean distance (unit: pixel/millimeter) between the reprojected tool tips and their ground truth as metrics for evaluation. The results are shown in Table~\ref{res2d}. Note that the original unit of this error is in pixels. We convert the pixel error to metric error (mm) by multiplying a scaling factor $s=\frac{Z}{f_x}$, where $Z$ and $f_x$ are the tool tip depth from triangulation and \textit{x}-axis focal length, respectively. As shown in Fig~\ref{fig_error}, we plot the 2D error versus frames to better analyze the error distribution.
\begin{table}[h]
\caption{Tool Tips 2D Reprojection Error (px/mm), L - Left, R - Right.}
\label{res2d}
\centering
\setlength{\tabcolsep}{2.5pt}
\renewcommand{\arraystretch}{1.15}
\begin{tabular}{lcccccc}
\hline
\textbf{} & \textbf{Video 0} & \textbf{Video 1} & \textbf{Video 2} & \textbf{Video 3} & \textbf{Video 4} \\
\hline
\textit{Avg. Err.} (L)  & 9.74/1.52 & 15.47/2.33 & 15.04/2.38 & 8.64/1.71 & 12.31/2.44 \\
\textit{Mdn. Err.} (L)  & 9.45/1.44 & 17.08/2.44 & 12.94/2.05 & 8.57/1.71 & 12.00/2.42 \\
\hdashline
\textit{Avg. Err.} (R)  & 7.71/1.49 & 10.33/1.50 & 7.52/1.41 & 17.43/2.90 & 13.65/2.22 \\
\textit{Mdn. Err.} (R)  & 6.83/1.29 & 9.84/1.48 & 7.08/1.34 & 16.59/2.63 & 11.98/2.05 \\
\hline
\end{tabular}
\end{table}

Using the stereo calibration parameters in SurgPose, corresponding keypoints $\mathbf{x}_r$ and $\mathbf{x}_l$ are triangulated to recover their 3D positions $\mathbf{x}$ in the left camera frame:
\begin{equation}
\mathbf{x} =
\arg\min_{\mathbf{x}}
\left(
\|\mathbf{x}_l - \pi(\mathbf{P}_l \mathbf{x})\|^2 +
\|\mathbf{x}_r - \pi(\mathbf{P}_r \mathbf{x})\|^2
\right),
\end{equation}
where $\pi(\cdot)$ denotes the perspective projection that converts homogeneous coordinates to image coordinates.
$\mathbf{P}_l=\mathbf{K}_L[\mathbf{I}\mid\mathbf{0}]$ and $\mathbf{P}_r=\mathbf{K}_R[\mathbf{R}\mid\mathbf{t}]$ are the stereo projection matrices derived from calibration.

We assume this $\mathbf{x}$ is the ground truth for the tool tip position. 
Given the joint angles and the corrected 6D pose, we can derive $\hat{\mathbf{x}}$, the 3D positions of the tool tips in the camera frame, using forward kinematics. As shown in Table~\ref{res3d}, we adopt the tool tip 3D Euclidean distance error $\left\|\mathbf{x}-\hat{ \mathbf{x}}\right\|$ to evaluate the performance. 

\begin{table}[h]
\caption{Tool Tips 3D Euclidean Error (mm), L - Left, R - Right.}
\label{res3d}
\centering
\setlength{\tabcolsep}{2.5pt}
\renewcommand{\arraystretch}{1.15}
\begin{tabular}{lcccccc}
\hline
\textbf{} & \textbf{Video 0} & \textbf{Video 1} & \textbf{Video 2} & \textbf{Video 3} & \textbf{Video 4} \\
\hline
\textit{Avg. Err.} (L)  & 4.84 & 5.46 & 5.66 & 7.22 & 6.76 \\
\textit{Mdn. Err.} (L)  & 4.75 & 5.45 & 5.65 & 7.11 & 6.75 \\
\hdashline
\textit{Avg. Err.} (R)  & 6.53 & 4.51 & 4.57 & 4.19 & 4.56 \\
\textit{Mdn. Err.} (R)  & 6.53 & 4.55 & 4.57 & 4.30 & 4.71 \\
\hline
\end{tabular}
\end{table}



\section{LIMITATION \& DISCUSSION}
In da Vinci Classic Surgical System, the reported dVRK end-effector pose is w.r.t. the ECM tip coordinate system. According to equation (1), the formulation can be denoted as ${}^{c}\mathbf{T}_{ee}={}^{c}\mathbf{T}_{ecm}\cdot{}^{ecm}\mathbf{T}_{ee}$, and we aim to recover the camera to ECM transformation ${}^{c}\mathbf{T}_{ecm}$. In the dVRK convention, the ECM frame origin is located at the midpoint of the stereo camera baseline. A naive assumption is therefore that the camera translation is approximately half of the baseline ($\sim$2.5mm)~\cite{avinash2019pickup}. However, in practice this rough approximation leads to centimeter-level translation errors of the end-effector, and the reprojected end-effector may even fall outside the image frame.

While Gaussian Splatting (GS) has demonstrated strong potential for general robotic self-representation, its application to articulated surgical instruments remains in an early stage of development. Based on our empirical observations, several limitations of the current instrument GS representation persist and highlight important directions for future improvement: 1) \textit{Limited novel-view rendering fidelity}. Although the geometry-aware pretraining of Instrument-Splatting enables robust semantic mask rendering, the photorealistic quality of novel-view synthesis remains limited. 2) \textit{Single instrument category coverage}. The current framework is only validated on LND, because training an Instrument-Splatting model for a specific instrument requires its corresponding CAD model. Thus, developing CAD-free GS representations across multiple instrument categories is a promising direction for future work. 3) \textit{Absence of explicit lighting modeling}. The present formulation does not explicitly model illumination, which is a major source of domain discrepancy between rendered images and real observations. Incorporating lighting-aware rendering or appearance adaptation may further reduce this gap and improve robustness.

In addition to the limitations of the GS representation, several sources of uncertainty are not explicitly modeled in the current framework.
The optical system of the modern endoscopic camera is highly complex. Kalia \textit{et al.}~\cite{kalia2020evaluation} suggest that the depth of the effective focal plane may vary with the surgical operative distance, i.e., the instrument depth. Furthermore, the classic Brown-Conrady model is insufficient to represent the intricate distortion of the endoscopic lens stack. Inspired by advances in neural lens modeling~\cite{xian2023neural, deng2025self}, integrating a learnable camera model into the GS-based differentiable rendering pipeline to jointly optimize the camera intrinsics and distortion is an attractive direction for future work. 
As shown in Fig.~\ref{fig_error}, the reprojection error exhibits temporal variation, suggesting a potential correlation with the kinematic parameters. In the proposed framework, we do not model this correlation. Incorporating a lightweight learning module to capture such correlations might further enhance the hand-eye calibration accuracy.


\section{CONCLUSION} We present SurgCalib, an automatic and markerless hand-eye calibration framework tailored for the da Vinci surgical system. To the best of our knowledge, this work is the first exploration of leveraging Gaussian Splatting (GS) representations for hand-eye calibration in surgical robotics. Experimental evaluation on the public dVRK SurgPose demonstrates its effectiveness in achieving accurate 2D reprojection and 3D tool-tip localization.









\bibliographystyle{IEEEtran}
\bibliography{references.bib}
\end{document}